\documentclass[sigconf]{acmart}
\usepackage[draft=true]{minted}
\usepackage{bbm}
\usepackage{xcolor}
\usepackage{algorithm,algpseudocode}
\algnewcommand{\Inputs}[1]{%
  \State \textbf{Inputs:}
  \Statex \hspace*{\algorithmicindent}\parbox[t]{.8\linewidth}{\raggedright #1}
}
\algnewcommand{\Initialize}[1]{%
  \State \textbf{Initialize:}
  \Statex \hspace*{\algorithmicindent}\parbox[t]{.8\linewidth}{\raggedright #1}
}
\usepackage{subfigure}
\usepackage[export]{adjustbox}
\newcommand{\algo}{All-Season }

\renewcommand\footnotetextcopyrightpermission[1]{} 
\AtBeginDocument{%
  \providecommand\BibTeX{{%
      \normalfont B\kern-0.5em{\scshape i\kern-0.25em b}\kern-0.8em\TeX}}}

\acmConference{}
\acmPrice{}
\acmISBN{}

\begin{document}

\title{A Linear Bandit for Seasonal Environments}

\author{Giuseppe Di Benedetto}
\email{benedett@stats.ox.ac.uk}
\affiliation{University of Oxford}
\authornote{This work has been done during an internship at the Amazon Development Centre in Berlin, Germany.}

\author{Vito Bellini}
\email{vitob@amazon.com}
\affiliation{Amazon, Berlin}

\author{Giovanni Zappella}
\email{zappella@amazon.de}
\affiliation{Amazon, Berlin}



\begin{abstract}
  Contextual bandit algorithms are extremely popular and widely 
  used in recommendation systems to provide online personalized recommendations.
  A recurrent assumption is the stationarity of the reward function, 
  which is rather unrealistic in most of the real-world applications. 
  In the music recommendation scenario for instance, people's music 
  taste can abruptly change during certain events, such as Halloween or 
  Christmas, and revert to the previous music taste soon after. 
  We would therefore need an algorithm which can promptly react to 
  these changes. Moreover, we would like to leverage already observed
  rewards collected during different 
  stationary periods which can potentially 
  reoccur, without the need of restarting the learning process from scratch. 
  A growing literature has addressed the problem of reward's 
  non-stationarity, providing algorithms that
  could quickly adapt to the changing environment. However, 
  up to our knowledge, there is no algorithm which deals with seasonal 
  changes of the reward function.
  Here we present a contextual bandit algorithm which detects and 
  adapts to abrupt changes of the reward function and leverages 
  previous estimations whenever the environment falls back to a 
  previously observed state.
  We show that the proposed method can outperform state-of-the-art algorithms
  for non-stationary environments.
  We ran our experiment on both synthetic and real datasets.
\end{abstract}






\maketitle

\section{Introduction}\label{intro}
Bandit algorithms are extremely popular and widely used in recommender systems, due to their ability to efficiently deal with the exploration-exploitation trade-off in an online fashion. Moreover, contextual bandits are able to leverage context information (e.g., regarding the user or the device) often available in modern applications.
\begin{figure}
  \centering
  \includegraphics[width=0.9\columnwidth, height=4cm]{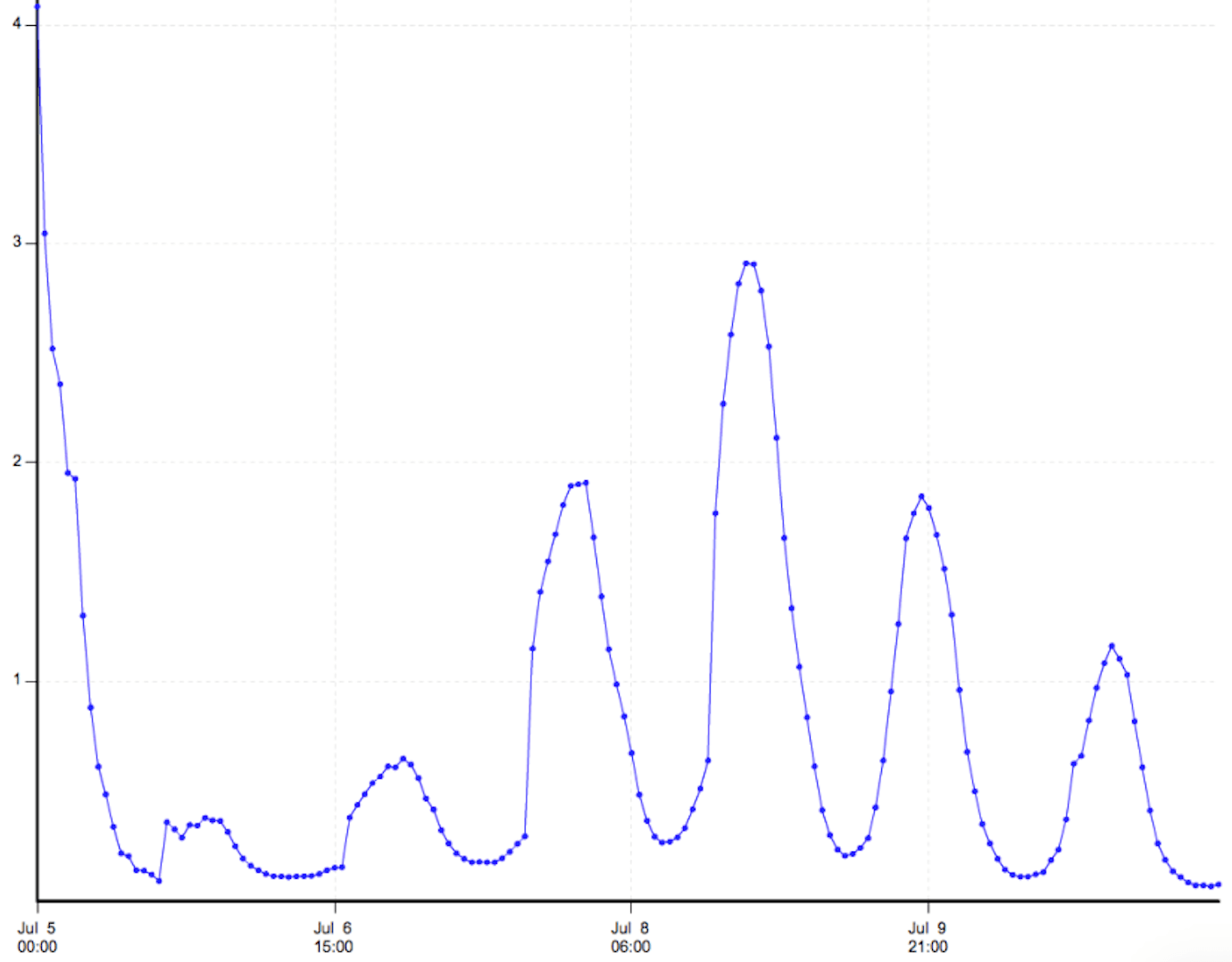}
  \caption{Average position over 5 minute time units on the page (lower positions are higher in the page) of a pre-selected marketing campaign for a large-scale music streaming service. Position is selected by a baseline method which can partially adapt over time.}
  \label{fig:musicdata}
\end{figure}
A common assumption which is often postulated is the stationarity of the environment. Under this assumption, the aim of the algorithm is to obtain a precise estimate of the parameter that defines the mapping between contextualized actions and rewards. However, most of the times this assumption is not satified in real-world scenarios. For example, in Figure~\ref{fig:musicdata} we report the average position assigned by a simple time-adapting algorithm to a musical item associated with a marketing campaign on a large-scale music streaming platform.
As we can observe, the taste of the users changes significantly over time and intra-day patterns are clearly visible, with some contents being extremely successful at certain hours of the day but performing significantly worse in other moments. It is also important to notice that while the patter is clearly visible, heuristics estimating the change points in the first few hours (or days) would not provide a reliable information in this case. Change points are visible but it is extremely hard to correctly estimate
them a-priori and for this reasons it is important to consider them unknown. 

This is just one example, which in practice is often addressed introducing additional information in the context. While these solution are often effective for well-known scenarios, they often fail in real-world recommenders due to the difficulty of keeping track of specific information about the environment. When a recommendation system is serving customers' requests world-wide, it should have available information about specific days (e.g., local holidays which can be different even within the same country), special offers for each location (e.g., ad-hoc marketing campaigns), endogenous events (e.g., marketing campaigns on TV and other media, music festivals) and many other factors. This is just unfeasible in practice, and even if not unfeasible, it would be hard to learn the effect of each of these factors since some of them are quite rare but still extremely valuable from a business prospective.

The ubiquity of non-stationary data has lead, in the recent years, to several works (e.g., see \cite{luo2017efficient}, \cite{karnin2016multi}, \cite{chen2019new}) about online learning with bandit feedback under non-stationarity. Though, when dealing with abruptly changing environments, it is reasonable to assume that the stationary configurations can reoccur over time. Up to our knowledge however, there is no work proposing a strategy which exploits the knowledge gained about past stationary states of the environment, with the need to pay extra exploration cost whenever a change-point occurs.

We propose a novel contextual bandit algorithm that detects abrupt changes and leverages the seasonality of the reward function. It deploys a collection of base bandit instances, each one learning about one of the unique reward stationary states. Each bandit gets assigned a weight reflecting how likely recent observations are to come from the stationary periods described by it. A short-term memory bandit is used to detect reward shifts toward stationary periods that had not occurred in the past, and cannot therefore be described by the existing bandits. In this case, a new base bandit instance is initialized. Experiments on real datasets show that our model can outperform state-of-the-art algorithms. 

The paper is organized as follows: Section~\ref{relwork} reviews some of the relevant algorithms present in the literature; Section~\ref{setting} details the setting and Section~\ref{algosection} contains a description of the proposed algorithm. Sections~\ref{expsynt} and ~\ref{expreal} show experiments on synthethic and real data, comparing our algorithm against several baselines. Section~\ref{conclusion} contains concluding remarks.

\section{Related work}\label{relwork}

In the last decade there has been an increasing interest in relaxing the stationarity assumption in the multi-armed bandits algorithms. While this is a step forward compared to the traditional stationary assumption, as already explained in Section~\ref{intro}, more and more complex scenario arises in real-world applications, and seasonal environments is just one of them.
The literature considering more complex scenarios is quite scarce in the bandit domain, so in this section we provide an overview of work studied for related settings.

In the non-stationary bandit problem, a key feature of the algorithms is to forget about past observations which could be outdated and therefore potentially harmful for current predictions. In this setting we can distinguish two main families of algorithms: the passive and the active.

Algorithms passively discarding the past do not perform any data-driven decision to choose what to discard and often rely on hyperparameters to decide how aggressively to forget.
Garivier and Moulines \cite{Garivier2011OnUB} first proposed two non-contextual mulit-armed bandit algorithms: the Sliding Window Upper Confidence Bound (SW-UCB) and the Discounted Upper Confidence Bound (D-UCB). The former simply updates the reward parameter estimator by considering a fixed number of past observations, while the latter discounts the past observations with exponentially decreasing weights, with the advantage that there is no need to store a sliding window of records to perform the update of the estimator. A recent work by Russac et al. \cite{russac2019weighted} extends the D-UCB to the linear case. Among the algorithms that passively address the non-stationarity, Besbes et al. \cite{NIPS2014_5378} propose a bandit-over-bandit schema leveraging the Exp3 algorithm~\cite{auer2002nonstochastic} with restart after a fixed and arbitrary number of observations, providing theoretical gurantees for this policy.
Allesiardo et al. \cite{Allesiardo2017} instead present a variation of the \emph{Successive Elimination with Randomized Round-Robin}~\cite{even2006action}, called \emph{Successive Elimination with Randomized Round-Robin and Resets} (SER3), where the estimators of the SER3 algorithm get reseted with a fixed probability to capture the best-arm switching taking place in changing environments.

Non-stationarity can be alternatively dealt with in an active fashion, by detecting when a change in the reward function occurs and propose a strategy to quickly adapt to the new environment. In the recent years several authors focused on this approach.
In the non-contextual bandit setting, Cao et al. \cite{Cao2019NearlyOA} present the \emph{Monitored-UCB} algorithm, a UCB instance with a change detector algorithm.
Allesiardo et al. \cite{Allesiardo2017} propose a variation of the Exp3  algorithm with a drift detection test.
Mellor and Shapiro \cite{mellor2013thompson} follow a Bayesian approach to tackle abrupt reward changes by linking the Thompson Sampling algorithm~\cite{thompson1933likelihood} with the Bayesian Online Change-point Detection (BOCPD).
These algorithms are all designed for the non-contextual bandit setting, so can hardly be leveraged in a real-world application and extending them to contextual bandit algorithms is often non-trivial. 

Within the class of bandit algorithms that actively detect changes in the enviroment, there are a few methods proposing to involve multiple multi-armed bandits in a hierarchical way.
One of the first was \emph{Adapt-EvE}, proposed by Hartland et al. \cite{Hartland2006MultiarmedBD}, which adopts a UCB instance together with the Page-Hinkley statistics to test for an abrupt change. A so called Meta-Bandit decides whether to accept the change detection or not, and the parameter of the change detection test is adjusted over time.
The method described by Cheung et al.\cite{bob}, called \emph{Bandit over Bandits} (BoB), involves a SW-UCB bandit whose window size is selected by a Exp3 bandit.
Wu et al. \cite{masterslaves} adopt a hierarchical bandit algorithm, called \emph{Dynamic Linear UCB} (dLinUCB), where a \emph{Master Bandit} decides which of the \emph{Slave Bandits} has to play based on the accuracy of the Slave bandits' reward estimations. Moreover, the Master bandit can discard out-of-date Slave bandits and create new ones.
A recent development of this algorithm, \emph{Dynamic Ensemble of Bandit Experts} (DenBand), is proposed by the same authors in \cite{denband}, and addresses context dependent reward changes.
DenBand is also suitable for the seasonal bandit scenario and we consider it as our main competitor in the experimental section.
An additional mention should be made about the work in \cite{adamskiy2012putting}.
While the algorithms described in that paper do not fit our setting but work in the prediction with expert advice setting, leveraging the full information feedback, it provides some interesting concepts, such as the mixing of past posteriors, that resemble our use of a collection of base bandits to make predictions (see Section~\ref{algosection} for details).

\section{Setting}\label{setting}
We present a novel contextual bandit algorithm for non-stationary reward functions with seasonality. In particular, the focus is on settings where the reward function abruptly changes, shifting to a brand new one or reverting to an already observed configuration. 
We assume that both the number of points in which the reward function changes (later called change-points) and the number of unique reward functions are unknown.
More in detail, we assume that at each time $t$ there is a set of possible actions to chose from $\mathcal{A}_t = \{x_{1}(t), \dots, x_{n_t}(t)\}$. In a music recommender system for example, an action could correspond to a track to be proposed to the user. Each action is represented by a contextualized action vector $x_i(t)\in\mathbb{R}^d$ which contains information about the context at time $t$ and about the action $i$, which in the music recommendation example can incorporate information about the user, the device, music genre of the track, etc.
The aim of a contextual bandit algorithm is to choose the action which maximizes the expected reward in a sequential fashion. In the ``seasonal setting'' we consider, the reward function that the bandit tries to learn is not unique but changes over time.
Given a sequence of steps from 1 to T, there are a number of so-called stationary periods $S=\{s_0, \dots, s_C\}$ in which a single reward function is used. Namely we assume that there are $C$ change-points determining stationary periods as a set of consecutive time steps. We assume $C \ll T$ and the change-points to be unknown to the learner, which therefore has to detect the change-points in order to quickly react, learning the parameters of the new reward function or selecting the appropriate one among the ones which it previously learnt.
Moreover, we assume a linear structure in the function of the rewards as follows
\begin{equation}\label{eq:reward}
  r_t(x_i(t)) = \langle \theta^*_t, x_i(t) \rangle + \epsilon_t
\end{equation}
where $\theta^*_t\in\mathbb{R}^d$ is the parameter of interest. Hereafter we assume the perturbations to be sampled from independent Gaussian distrubutions $\epsilon_t \stackrel{iid}{\sim}\mathcal{N}(0,\sigma^2)$. 

Under the seasonal assumption described above, the true value of the parameter $\theta^*$ is not constant over time, but it is selected from the set $\Theta = \{ \widetilde\theta_{1}, \dots, \widetilde\theta_{k} \}$ and each of the stationary periods $s\in S$ is associated with a unique value of the reward parameter $\theta^*_t = \widetilde\theta_s\in \Theta$ for all $t \in s$. It is also to be noted that in practice most often $k \ll C$.

\section{All-season bandit}\label{algosection}
The algorithm we propose addresses non-stationary rewards with abrupt and possibly seasonal changes. The seasonality of the reward function can be favourable since predictions about recurrent stationary states can be reused and further updated. This suggests the use of a collection of contextual bandits, each one targeting a stationary state of the reward function. The number of change-points is of course unknown a priori, therefore the algorithm has to detect these changes and recognize whether such a state had already occurred. In such case, the bandit which had learned about that value of the reward parameter needs to be chosen to interact with the environment. Alternatively, a brand new bandit needs to be created.
In order to perform the change-point detection, at each step every bandit gets assigned a score indicating how likely the last observation is to come from one of the stationary periods represented by the available bandits. The normalized scores are used to sample the bandit which will play in the current time step (see line~\ref{state:select} in algorithm~\ref{alg:vivaldi}). The collection of bandits used in the algorithm consists of a set of ``long-term memory'' instances, which are called \emph{base bandits} (denoted by $L_i$ in algorithm~\ref{alg:vivaldi}), and a ``short-term memory'' one (denoted by $S$ in algorithm~\ref{alg:vivaldi}), which is referred to as \emph{shadow bandit}. 
Each base bandit has to learn about one of the true values $\theta^*_s\in\Theta$ of the reward parameter. The shadow bandit, which only considers the most recent observations, is responsible for triggering the detection of changes to stationary states which had not been previously observed.

In real-world applications, it is common to receive observations gathered into batches with potentially inhomogeneous sizes. Recommender systems usually perform updates of their models according to a time schedule, and between each consecutive updates the system collects different numbers of feedbacks. Algorithm~\ref{alg:vivaldi} details the batch version of the proposed algorithm, with the shadow bandit being a sliding-window bandit instance.
Three hyperparameters are given as input: $\lambda$ is a regularization parameter which depends on the bandit prototype used for the base and shadow instances; $\tau$ tunes the short-term memory of the shadow bandit\footnote{The parameter $\tau$ could be the window size or the discount factor if the shadow bandit is a sliding-window or discounted bandit instance respectively}; $N_{\text{max}}$ is the maximum number of base bandits the learner is willing to maintain.
We denote with $B$ the number of batches. For each observation in the current batch, a bandit is sampled from the set of available bandits (line~\ref{state:select} in Algorithm~\ref{alg:vivaldi}) with probabilities proportional to the weights $\omega_j$, and interacts with the environment. If the shadow bandit has been selected at least once in the batch, then a new long-term memory instance is initialized with the parameters of the shadow bandit.
In the update step, the shadow bandit gets updated with the last $\tau$ observation, regardless of which bandit has played, while the base instances are updated on the set of observations they have been assigned in the batch. A maximum number $N_{\text{max}}$ of base bandits can be fixed a priori. If the number of base bandits exceeds this constraint, a pruning scheme is called to discard one of them. Before starting a new batch, the weights of each bandit are computed.

\subsubsection*{Reward shift detection}
The changes in the reward function are detected by looking at the weights assigned to each bandit, which depend on the last observed reward ---or the rewards observed in the last batch--- and on the reward parameter estimate provided by the bandit. 
When the reward function shifts to an already observed state, the base bandit which had learned about that stationary configuration should get assigned a high weight. Instead, when the reward parameter shifts to a new state which had not been previously observed, none of the base bandits would be eligible to choose the action to play. In this case a brand new bandit should be initialized. The creation of a new bandit is triggered by the shadow bandit. Being it a short-term memory instance, it does not take into account the whole history but, in the case of the sliding window base algorith, a fixed number $\tau$ of past observations (other base algorithms will adopt a different way to ``forget'' the past). Moreover, the shadow bandit gets updated at each time step, regardless of which bandit is playing (see line~\ref{state:supdate} of Algorithm~\ref{alg:vivaldi}), allowing it to track changes in the environment.
When the shadow bandit is selected to play, the new base instance is initialized as a copy of the shadow bandit (see line~\ref{state:shadowcopy} of algorithm~\ref{alg:vivaldi}).
diction are constantly based on the last $\tau$ observations. 

\begin{algorithm}
  \caption{All-Season bandit (batch update)}
  \label{alg:vivaldi}
  \begin{algorithmic}[1]
    \Inputs{$\lambda,\, \tau,\, N_{\text{max}}$}
    \Initialize{
      Create base and shadow bandits: $L_1, S=$ Init($\emptyset$)\\
      Set of base bandits: $\mathcal{B} = \{L_1\}$, $\;N_L=1$\\
      Initialize bandits' weights: $\omega_S = 0.5,\, \omega_1 = 0.5$
    }\label{state:init}
    \For{b = 1 to B}
    \State Set $I_j = \emptyset$ for all $j \in \mathcal{B} \cup \{S\}$
    \For{t in b}
    \State Bandit selection: $j\sim\text{Discrete}(\omega_j:j\in\mathcal{B}\cup\{S\})$
    \label{state:select}
    \State $I_j = I_j \cup \{t\}$
    \State $x_t =$ PlayBandit($j$, $\mathcal{A}_t$)
    \State Observe reward $r_t$
    \EndFor
    \If{$I_S \neq \emptyset$}
    \State $N_L = N_L + 1$
    \State Create bandit $L_{N_L}=$ Init($S$), $\;I_{L_{N_L}} = \emptyset$\label{state:shadowcopy}
    \State $\mathcal{B} = \mathcal{B} \cup \{L_{N_L}\}$ \label{state:lmnew}
    \EndIf
    \State $S=$ Init($\emptyset$) and  Update($S$,  $(x_s, r_s)_{s=\min (t-\tau+1,0):t}$) \label{state:supdate}
    \For{$j\in \mathcal{B}$}
    \State Update base bandit $j$: Update($L_j$, $(x_s, r_s)_{s\in I_j}$) \label{state:lmupdate}
    \EndFor
    \If{$N_L > N_{\text{max}}$}
    \State Prune($\mathcal{B}$) (see alg.~\ref{alg:pruningKL}) \label{state:pruning}
    \EndIf
    \State $(\omega_j)_{j\in\mathcal{B}\cup \{S\}}=$ UpdateWeights($\mathcal{B}\cup \{S\}$, $(x_s, r_s)_{s\in b}$)\label{state:weights}
    \EndFor
  \end{algorithmic}
\end{algorithm}

\subsection{Shadow bandit and base bandits}

In algorithm~\ref{alg:vivaldi} the choice of the bandit instances and of the bandits' weighting strategy are not specified. The weights assigned to each bandit should reflect the likelihood of describing the last observation. Any contextual bandit algorithm which can provide such scores can be used as prototype instance for base and shadow bandits in the proposed scheme. In the Bayesian framework, such weights are naturally provided by the posterior predictive probabilities of the observation under each bandit. Therefore in what follows, all the bandits instances involved in the algorithm are linear Thompson Sampling (linTS) bandits. Algorithm~\ref{alg:lints} recalls the procedures involved in the linear Thompson Sampling.
A conjugate model is adopted in order to have analytical form of the posterior distribution of the reward parameter. In particular we place a Gaussian prior on the parameter of interest and assume the reward likelihood to be Gaussian:
\begin{align}\label{eq:model}
  \theta &\sim \mathcal{N}\left(0_d,\, \lambda^{-1}\mathbbm{1}_d\right)\\
  r_t\,|\,\theta, x &\sim \mathcal{N}\left(\langle \theta, x\rangle,\,\sigma^2\right) 
\end{align}
Denoting by $D_t=\{(x_s,r_s)\}_{s=1:t}$ the history of chosen actions $x_s$ and corresponding observed rewards $r_s$ up to time $t$, the posterior distribution of the reward parameter, given $D_t$, is a Gaussian distribution with precision matrix $M = \lambda\mathbbm{1}_d + \sum_{s=1}^t x_sx_s^T$ and mean vector $\mu = M^{-1} \sum_{s=1}^t r_sx_s$.
The posterior predictive distribution of the observed reward at time $t$ given the previous observations is Gaussian distributed as well
\begin{equation}\label{eq:postpred}
  r_t\,|\, x_t, D_{t-1} \sim \mathcal{N}\left(\langle \mu, x\rangle,\,
    \sigma^2 + x_t\,M^{-1}\, x_t^T\right)
\end{equation}
with $\mu$ and $M$ being the mean vector and precision matrix of the posterior distribution of $\theta\,|\,D_{t-1}$ respectively.

Every base linTS bandit gets updated only when it plays, therefore its posterior distribution takes into account only the observations that have been assingned to it. Let $\mathcal{I}$ be the set of time indices one of the base bandits has played overall. The posterior distribution of the reward parameter provided by that base bandit is a Gaussian distribution with precision matrix $M = \lambda\mathbbm{1}_d + \sum_{s\in\mathcal{I}}x_sx_s^T$ and mean vector $\mu = M^{-1}\sum_{s\in\mathcal{I}}r_sx_s$.

The changes in the reward function are detected by looking at the posterior predictive probabilities of the last observed reward. In the case the rewards are observed in batches, the scores assigned to each bandit take into account all the records in the last batch (see algorithm~\ref{alg:updateweights}). Denoting by $p_j(r_t\,|\,x_t)$ the posterior predictive probability that the bandit $j$ gives to the reward $r_t$, given the action $x_t$, we have that the weight assigned to the bandit after collecting a batch $b$ of observations is
\[
\omega_j = \prod_{t\in b}p_j(r_t\,|\,x_t)
\]
After computing these scores for each bandit, the algorithm samples the reward parameter estimate $\hat\theta$ from the mixture of posterior distributions provided by the available bandits, with weights being the predictive posterior probabilities defined above (line~\ref{state:select} of Algorithm~\ref{alg:vivaldi}).

\begin{algorithm}
  \caption{Linear Thompson Sampling}
  \label{alg:lints}
  \begin{algorithmic}[1]
    \Procedure{Init}{L}
    \If{L $=\emptyset$}
    \State \Return $M = \lambda \mathbbm{1}_d,\, b = 0_d,\, \mu = 0_d$
    \Else
    \State \Return $M = M_L,\, b = b_L,\, \mu = \mu_L$
    \EndIf
    \EndProcedure
\\\hrulefill
    \Procedure{PlayBandit}{L, $\mathcal{A}_t$}\label{state:playlints}
    \State Sample $\hat\theta_j \sim \mathcal{N}(\mu_L,\, M_L^{-1})$ \label{state:sample}
    \State \Return $x_t = \text{arg}\max_{x\in\mathcal{A}_t} \langle \hat\theta_j, x\rangle$ \label{state:bestarm}
    \EndProcedure
\\\hrulefill
    \Procedure{Update}{L, $(x_s, r_s)_{s\in I}$}
    \State $M_L = M_L+\sum_{i\in I}x_ix_i^T,\, b_L=b_L+\sum_{i\in I}x_ir_i ,\, \mu_L=M_L^{-1}b_L$
    \EndProcedure
  \end{algorithmic}
\end{algorithm}

\begin{algorithm}
  \caption{Bandit's weights update}
  \label{alg:updateweights}
  \begin{algorithmic}[1]
    \Procedure{UpdateWeights}{$\mathcal{I}$, $(x_s, r_s)_{s\in b}$}
    \State $\omega_j = \prod_{t\in b}p_j(r_t\,|\,x_t)$ for all $j\in\mathcal{I}$
    \State \Return $(\omega_j)_{j\in\mathcal{I}}$
    \EndProcedure
  \end{algorithmic}
\end{algorithm}

\subsection{Pruning schemes}
Allowing the algorithm to create an unbounded number of base bandits could lead to computational issues, both in terms of memory and time. As a matter of fact, outliers might trigger false change-point detections which would imply the creation of additional long-term memory bandits targeting a stationary configurations which had already been learnt by another long-term memory bandit previously created. The problem of having spurious base bandits can be addressed by imposing the maximum number of long-term memory instances we are willing to mantain. This number can be chosen using some prior information about the number of unique stationary configurations $|\Theta|$. However, in the case such knowledge is missing, it is recommended to be conservative about this constraint. Having multiple base bandits learning the same stationary period would result in a slow convergence of their estimators since they get updated less frequently.
In order to satisfy the constraint, the algorithm needs to have a strategy to prune the least useful base bandit whenever the maximum allowed number is exceeded. We propose two different pruning schemes to control the number of available bandits at each time step (see Algorithm ~\ref{alg:pruningKL}).
The rationale behind both schemes is to find the pair of closest bandits, and among the two, discard the one  which is less certain about its estimate, measured with the trace of the associated posterior covariance matrix. A quick way to find the pair of closest bandits would be to compare the pairwise distances between the posterior mean parameters of the base bandits, resulting in a time complexity of $\mathcal{O}(N_{\text{max}}^2d)$. However, this would not take into account the whole distribution associated to each bandit. Therefore an alternative approach, used in the experiments presented below, consists in comparing the symmetric Kullback-Leibler divergence between pairs of base bandits. The symmetric Kullback-Leibler divergence between two measures $p$ and $q$ is defined as $\text{KL}_{\text{sym}}(p, q) = \text{KL}(p,q) + \text{KL}(q, p)$, and it can be computed in closed form if the distributions of interest are Gaussian:
\begin{align}\label{eq:KL}
  &\text{KL}_{\text{sym}}(\mathcal{N}(\mu_1,\, \Sigma_1),\,\mathcal{N}(\mu_1,\,\Sigma_2))=
  \frac{1}{2} \text{tr}(\Sigma_2^{-1}\Sigma_1 + \Sigma_1^{-1}\Sigma_2)\\
  &+\frac{1}{2} (\mu_2-\mu_1)^T\Sigma_2^{-1}(\mu_2-\mu_1) +  (\mu_2-\mu_1)^T\Sigma_1^{-1}(\mu_2-\mu_1) \nonumber
\end{align}
The time complexity of this scheme is $\mathcal{O}(N_{\text{max}}^2d^3)$.

\begin{algorithm}
  \caption{KL pruning}
  \label{alg:pruningKL}
  \begin{algorithmic}[3]
    \Inputs{Long-memory bandits $j\in\mathcal{I}$}
    \State $(i^*, j^*) = \text{arg}\min_{(i,j) \in \mathcal{I}^2, i<j}
    \text{KL}_{\text{sym}}(\mathcal{N}(\mu_j,\, M_j^{-1}),\,\mathcal{N}(\mu_j,\, M_j^{-1}))$
    \State $i_{\text{delete}}=\text{arg}\min_{k = i^*,j^*}\text{tr}(M_{k}^{-1})$
    \State $\mathcal{I} = \mathcal{I} \setminus \{L_{i_{\text{delete}}}\}$ and $N_L = N_L - 1$
  \end{algorithmic}
\end{algorithm}

\section{Synthetic data experiment}\label{expsynt}
\begin{figure}[h!]
  \includegraphics[width=0.92\columnwidth, left, height=4cm]{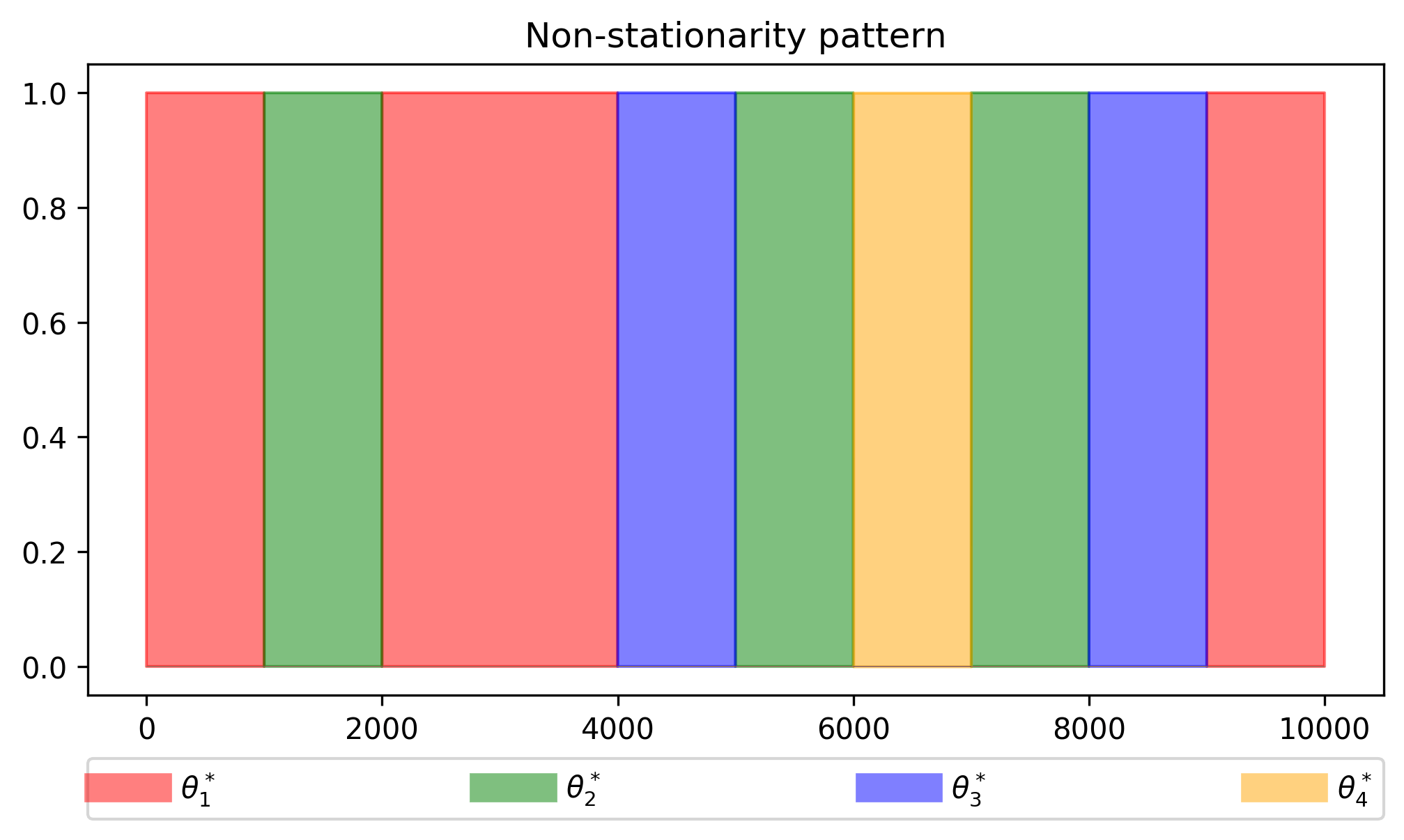} \hspace*{0.4cm}
  \includegraphics[width=\columnwidth, left, height=4cm]{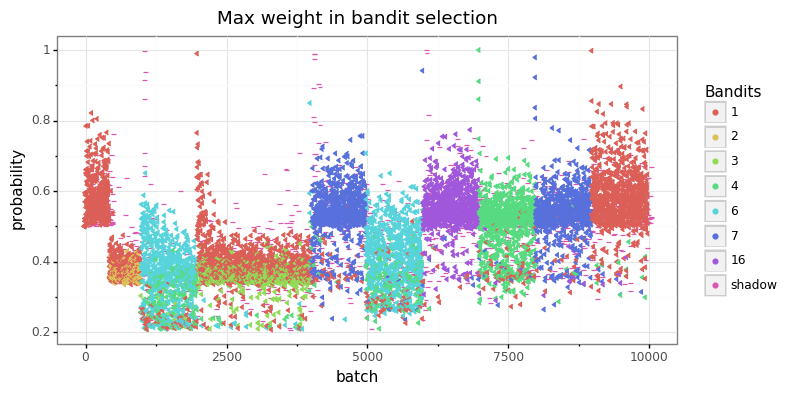}
  \caption{Top figure: Non-stationary pattern of the reward function. Bottom figure: weight of the bandit with highest posterior predictive weight, base bandits depicted with coloured triangles and shadow bandit with pink dashes.}
  \label{fig:simulated}
\end{figure}
The first experiment is on synthetic data with abruptly changing and seasonal rewards. The aim is to show empirically how the proposed model functions.
The experiment consists of $n=10000$ observations. At each time, the action set consists of five unit-norm vectors in $\mathbb{R}^5$ obtained by randomly sampling from a centred Gaussian distribution with independent coordinates and normalizing. The rewards are sampled from a Gaussian distribution as described in eq~\ref{eq:reward} with $\sigma^2=0.1$ and reward parameter $\theta^*\in\mathbb{R}^5$ changing over time. We consider four different values for the reward parameter, representing the unique stationary periods, alternating over time according to the pattern depicted in Figure~\ref{fig:simulated}, which contains eight change-points. In this example each batch corresponds to one observation, and the window-size of the Shadow bandit is set to $\tau=10$. Under correct specification of the model, the proposed algorithm manages to recover the stationary periods. The second plot in Figure~\ref{fig:simulated} shows the posterior predictive weights of the most likely bandit to be chosen at each time step. The base bandits indexed by 1, 2, 3 learn about the stationary period represented by $\theta^*_1$ (see first plot in Figure~\ref{fig:simulated}); $\theta^*_2$ is learned by base bandits 4 and 6; while $\theta^*_3$ and $\theta^*_4$ by bandits 7 and 16 respectively. It is worth noting that the shadow bandit has probability of being chosen very close to $1$ only when there is a change-point followed by a stationary period which had not been previously observed, namely in the first, third and fifth change-points. In contrast, when an alredy observed stationary period reoccurs, the long-term memory bandit which had been learning about it gets assigned a probability very close to $1$, meaning that the reward seasonality is correctly captured. Even if the shadow bandit is assigned the highest weight very often, the constraint on the maximum allowed number of long-term memory bandits ensures that the spurious long-term memory bandits, that were created when not necessary, are discarded by the pruning scheme.

\section{Experiments with real data}\label{expreal}
In this section we present a collection of experiments to compare the \algo bandit against non-stationary baselines (SW-LinTS, D-LinTS, BoB \cite{bob}, DenBand \cite{denband}), showing that our model can provide better performance on abruptly changing environments with seasonality.
Even though the best way to compare bandit algorithms would be to test them online with A/B tests, online experiments in recommedation systems are usually expensive and risky since they can negatively affect the customer experience. Moreover, they are impossible to reproduce and unaccessible to academic community. Methods to perform offline evaluation have been designed in order to compare different algortihms on randomized logged data (e.g., \cite{li2011unbiased}, \cite{li2010contextual}). The authors of \cite{dudik2012sample} propose an evaluation scheme for non-stationary policies, although the assumption of iid contexts is violated in our case.
Moreover, we are considering the case where both the logging policy and the environment are non-stationary, and up to our knowledge, there is not any unbiased offline evaluation proceduce already known for this setting. In order to offer a fair comparison, we have artificially induced non-stationarity to classification problems using two real datasets, trying to mimic plausible non-stationary patterns that could occur in real-world applications.

\subsection{Change-points patterns}
\begin{figure}[h!]
  \centering
  \includegraphics[width=0.9\columnwidth, height=4cm]{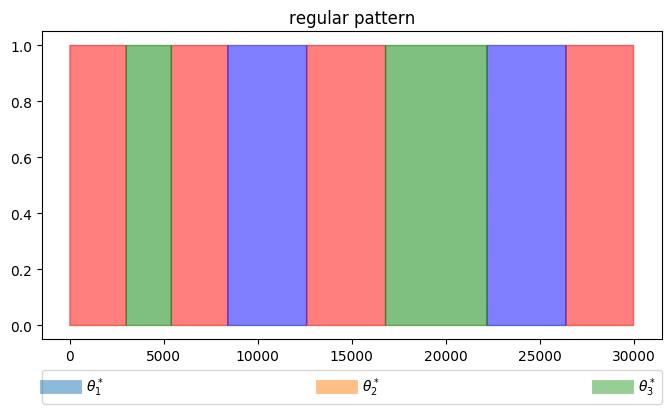}
  \includegraphics[width=0.9\columnwidth, height=4cm]{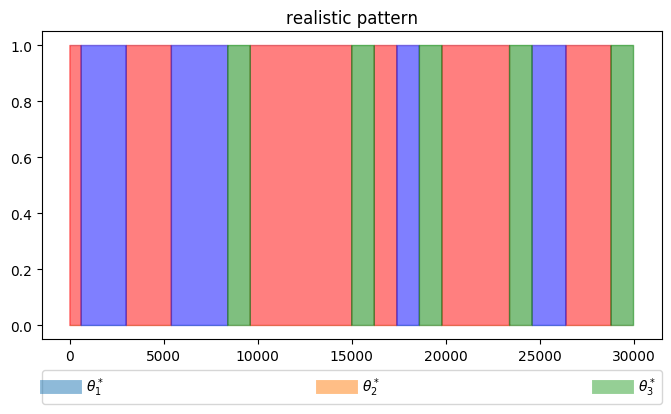}
  \includegraphics[width=0.9\columnwidth, height=4cm]{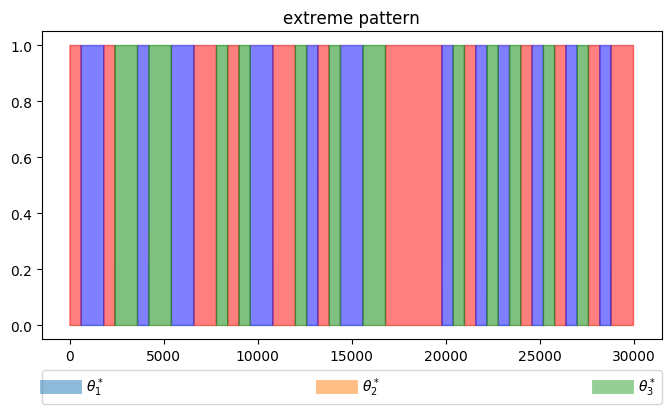}
  \caption{Non-stationary patterns}
  \label{fig:patterns}
\end{figure}

In the following experiments we compare non-stationary bandit algorithms on real datasets with three different types of reward seasonality. In all these settings the reward function is piece-wise constant, therefore the changes are abrupt, and the number of unique stationary configurations is the same across them. However, the number of change-points differs, and therefore the length of each stationary period (see Figure~\ref{fig:patterns}). The number of changes and their lenght are designed using real data from a large-scale music streaming service.
\subsubsection*{Regular pattern} The first scenario is characterized by long stationary periods and few change-points. This setting should favour the algorithms which passively address non-stationarity, such as those which discount past observations or base their prediction on a sliding window of records. Although such algorithms are not suited for abrupt changes, the long duration of a stationary period allows them to learn about the reward parameter before encountering the following change-point;
\subsubsection*{Realistic pattern} The second scenario contains a mix of short and long stationary periods. The aim is to mimic abrupt-changes which might occur in real-world scenarios. In the music recommendation paradigm, very short stationary periods could represent the occurrence of short events (Halloween or Father's Day for instance) which abruptly change the music taste of the users for a fairly short time. Longer stationary periods could correspond for instance to Summer or Christmas, when users might be more willing to listen to Summer hits and Christmas songs respectively. Very long periods instead could correspond to the usual music taste of the user, without being affected by external events;
\subsubsection*{Extreme pattern} All the stationary periods are very short. This scenario can occur when looking at a finer time grid, describing for example variation in the music listened during daytime and nighttime (see for example Figure~\ref{fig:musicdata}).
For the SW-linTS the learner should pick a window wider than the learning time of the LinTS, since this would ensure a good performance in the stationary setting. However, it might be the case that the stationary periods are much shorter than the convergence time. In such case, the bandits that passively tackle the non-stationarity, e.g. SW-linTS and D-linTS, would not be able to quickly react and converge to such short stationarity periods. 

\subsection{Datasets}
\begin{figure}
  \includegraphics[width=0.9\columnwidth, height=4cm]{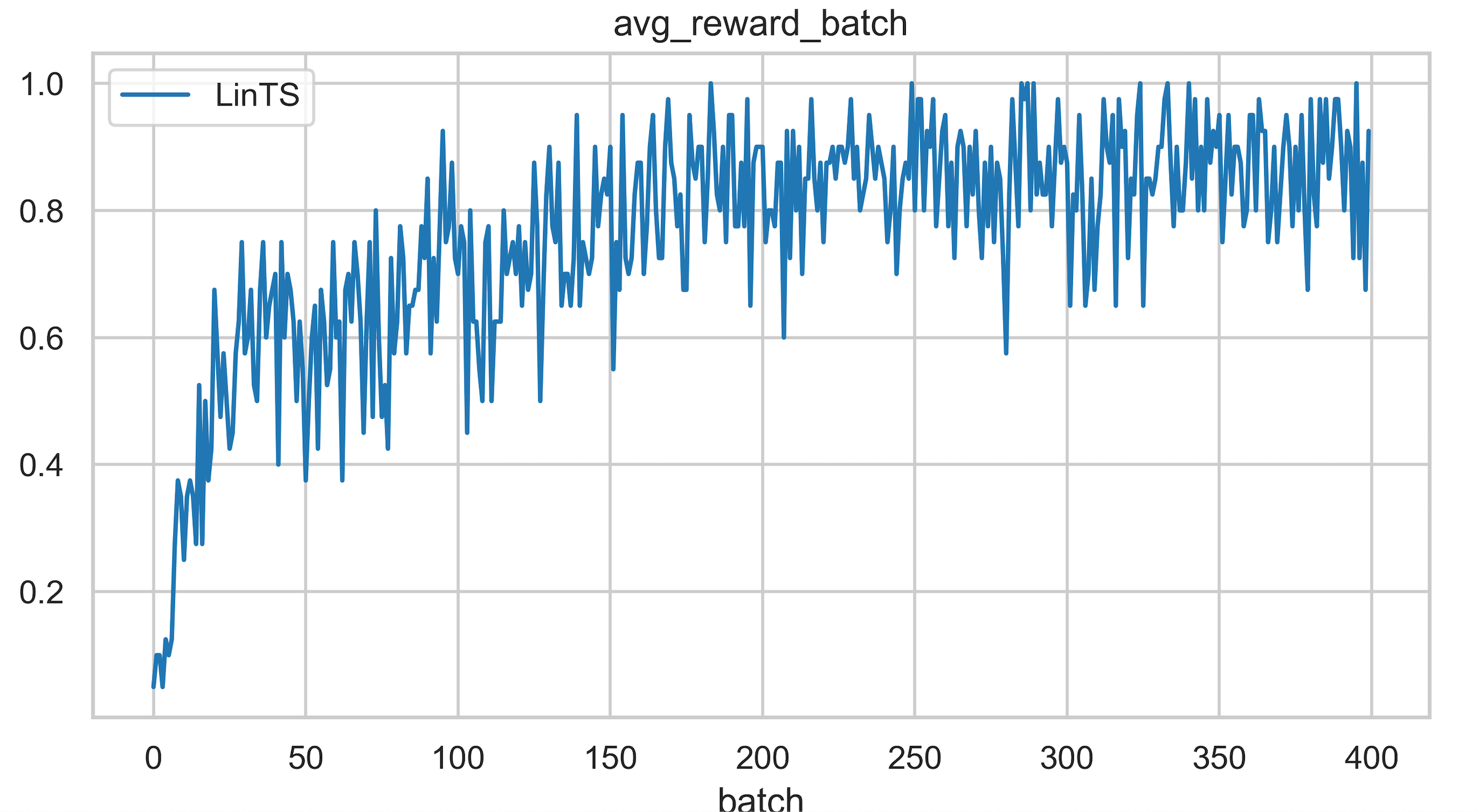}
  \caption{LinTS average reward per batch in the MNIST digit recognition task. Each batch contains 10 observations.}
  \label{fig:lints_LT}
\end{figure}

Two real datasets are considered in the following experiments: MNIST and Fashion-MNIST. Both datasets are used for benchmarking algorithms on image classification. Each observation consists of a 784-dimensional vector with entries from $0$ to $255$ representing the pixel values of a grey-scale image of a digit (MNIST) or a fashion item (Fashion-MNIST).
PCA is perfomed to reduce the dimensionality of the context vectors down to $44$ principal components for MNIST and $43$ for Fashion-MNIST, describing $80\%$ and $85\%$ of the variance respectively.
Across all the experiments we run all the algorithms on the first $30000$ observations gathered into batches of size $10$. Each stationary period lasts at least for $600$ observations and the convergence time of the linTS on the MNIST digit recognition task is of about $2000$ observations (see Figure~\ref{fig:lints_LT}). The contextualized action vectors are given by the outer product between the one-hot encoded action vector and the context vector containing the image features. Two different experiments are presented for each dataset.

\subsubsection*{Two-arm experiment}
The first setting consists of a two-arm problem. In the MNIST dataset, three different tasks are considered: parity, divisibility by three, and primality of the digit. In particular the the two arms $($arm 1, arm 2$)$ represent the actions $($even, odd$)$, $($divisible by three, non divisible by three$)$, $($non prime, prime$)$ for each task respectively. Reward $1$ is given for correct classification, and $0$ otherwise. Analogously for Fashion-MNIST, the arms describe the following classification tasks: $($upper-body, lower-body$)$, $($winter clothes, summer clothes$)$, $($shoes, clothes$)$.

\subsubsection*{Arm-shift experiment}
In this setting the standard classification of the digits and fashion items is performed. The non-stationarity is induced by shuffling the labels of the arms, therefore each stationary period is represented by a permutation of the labels.

\subsection{Baselines and hyperparameters' setting}
Here we list the baselines used in the experiments, with the relative hyperparameters' setting. Hyperparameter optimization is performed by grid search using the first $10\%$ of the data as validation set (the initial part of the dataset). The regularization hyperparameter is present in all the baselines, hence it has been set to $\lambda=1$ in all the experiments. 
\begin{itemize}
  \item[SW-linTS]: Sliding-window LinTS with window size $\tau$ optimized in the set $\{50,\,100,\,500,\,1000,\,5000\}$;
  \item[D-linTS]: Discounted LinTS, the discount parameter \mbox{$\gamma=1-10^{-\kappa}$} with $\kappa\in\{1, 3, 5, 10\}$;
  \item[BoB]: \emph{Bandit Over Bandit} algorithm presented in \cite{bob}
    The time horizon is provided a priori, since it is needed to define the set of allowed window-sizes according to equation (23) in \cite{bob}. We optimize for the hyperparameters denoted by $S$ and $L$ in the paper \cite{bob} in the set $\{0.1,\, 1,\, 10\}$;
  \item[DenBand]: \emph{Dynamic Ensamble of Bandit} (\cite{denband}) can be considered as the main competitor of our algorithm, since it uses a collection of bandits to address non-stationarity and in particular context-dependent reward changes. A sliding window size $\tau$ described in the paper is optimized in the set $\{500,\, 1000,\, 2000\}$, while the hyperparameter $\Delta_L$ in the set $\{0.1,\, 0.25,\, 0.5\}$. 
  \item[LinTS]: Linear Thompson Sampling, for which no hyperparameter optimization is needed, this baseline is included to have a comparison against a bandit algorithm which does not take into account non-stationarity; 
  \item[Random]: picks an action uniformly at random.
\end{itemize}
\algo bandit has two hyperparameters: the maximum allowed number $N_{\text{max}}$ of base bandits and the window-size $\tau$ of the shadow bandit. They are optimized over the sets $N_{\text{max}}\in\{3,\, 4,\, 5\}$ and $\tau\in\{50,\, 100,\, 500,\, 1000,\, 5000\}$.

\subsection{Results on two-arms experiment}
In the first experiment with only two arms the reward function partially changes between two distinct stationary periods. For instance, in the MNIST dataset, when the task switches from classifying the parity to classifying the divibility by three, the best arm does not change for the context vector representing the digit six. In both datasets \algo bandit (both versions) are almost always better than the base algorithms employed, with a single dataset in which there is a substantial parity. Moreover, we see that \algo(Disc) is mostly 
outperforming the \algo(SW). In addition, both the SW-LinTS and D-linTS outperform the other more complex baselines BoB and DenBand on both datasets.
\subsection{Results on experiments with arm shifts}
The second experiment is a ten-arm bandit problem, and in this case the reward parameter changes completely between two distinct stationary periods. The \algo bandits constantly ouperform 
all the competitors, in some case with large margin. Generally, SW-LinTS is the best baseline followed by the D-linTS, while the remaining baselines are often underperforming by large margin.

\begin{table}[h!]
  \centering
  \begin{tabular}{rccc}
    \toprule
    & Regular & Realistic & Extreme \\
    \midrule
    \algo(Disc)  & 0.77 $\pm$ 0.01 & 0.74 $\pm$ 0.02 & 0.77 $\pm$ 0.01  \\
    \algo(SW)  & 0.77 $\pm$ 0.01 & 0.78 $\pm$ 0.01 & 0.74 $\pm$ 0.01  \\
    SW-LinTS  & 0.76 $\pm$ 0.00 & 0.76 $\pm$ 0.00 & 0.71 $\pm$ 0.00 \\
    D-LinTS   & 0.74 $\pm$ 0.00 & 0.73 $\pm$ 0.00 & 0.71 $\pm$ 0.00 \\
    DenBand   & 0.71 $\pm$ 0.00 & 0.71 $\pm$ 0.00 & 0.69 $\pm$ 0.00 \\
    BoB       & 0.65 $\pm$ 0.07 & 0.56 $\pm$ 0.08 & 0.69 $\pm$ 0.06 \\
    LinTS     & 0.72 $\pm$ 0.00 & 0.71 $\pm$ 0.00 & 0.70 $\pm$ 0.00 \\               
    random    & 0.50 $\pm$ 0.00 & 0.50 $\pm$ 0.00 & 0.50 $\pm$ 0.00 \\
    \bottomrule
  \end{tabular}
  \caption{\small Mnist two-arm: average reward $\pm$ std dev.}
  \label{table:mnist2}
\end{table}

\begin{table}[h!]
  \centering
  \begin{tabular}{rccc}
    \toprule
    & Regular & Realistic & Extreme \\
    \midrule
    \algo(Disc)  & 0.81 $\pm$ 0.02 & 0.81 $\pm$ 0.01 & 0.81 $\pm$ 0.00  \\
    \algo(SW)  & 0.80 $\pm$ 0.02  & 0.79 $\pm$ 0.01  & 0.81 $\pm$ 0.02   \\
    SW-LinTS   & 0.77 $\pm$ 0.00  & 0.77 $\pm$ 0.00 & 0.76 $\pm$ 0.00   \\
    D-LinTS    & 0.75 $\pm$  0.00 & 0.68 $\pm$ 0.00 & 0.72 $\pm$ 0.00   \\
    DenBand    & 0.65 $\pm$ 0.01 & 0.64 $\pm$ 0.01 & 0.64 $\pm$ 0.01  \\
    BoB        & 0.68 $\pm$ 0.00 & 0.68 $\pm$ 0.00  & 0.70  $\pm$  0.00 \\
    LinTS      & 0.67 $\pm$ 0.01 & 0.68 $\pm$  0.00 & 0.69 $\pm$  0.00  \\               
    random     & 0.50 $\pm$ 0.01 & 0.50 $\pm$ 0.01 & 0.50 $\pm$  0.01 \\
    \bottomrule
  \end{tabular}
  \caption{\small Fashion-Mnist two-arm: average reward $\pm$ std dev.}
  \label{table:fmnist2}
\end{table}

\section{Conclusion and future work}\label{conclusion}
We have proposed a new bandit algorithm which addresses the problem of learning in non-stationary environments. Our algorithm
can employ a large class of potential base algorithms (in our case two variants of Linear Thompson Sampling) and in our experimental results it always outperformed the corresponding base algorithms.
The algorithm proved itself able to adapt to different non-stationarity patterns and outperform state-of-the-art solutions by large margin.
Moreover, \algo is a significantly 
\clearpage
\begin{figure*}[h!]
\begin{tabular}{ccc}
  \includegraphics[scale=.30]{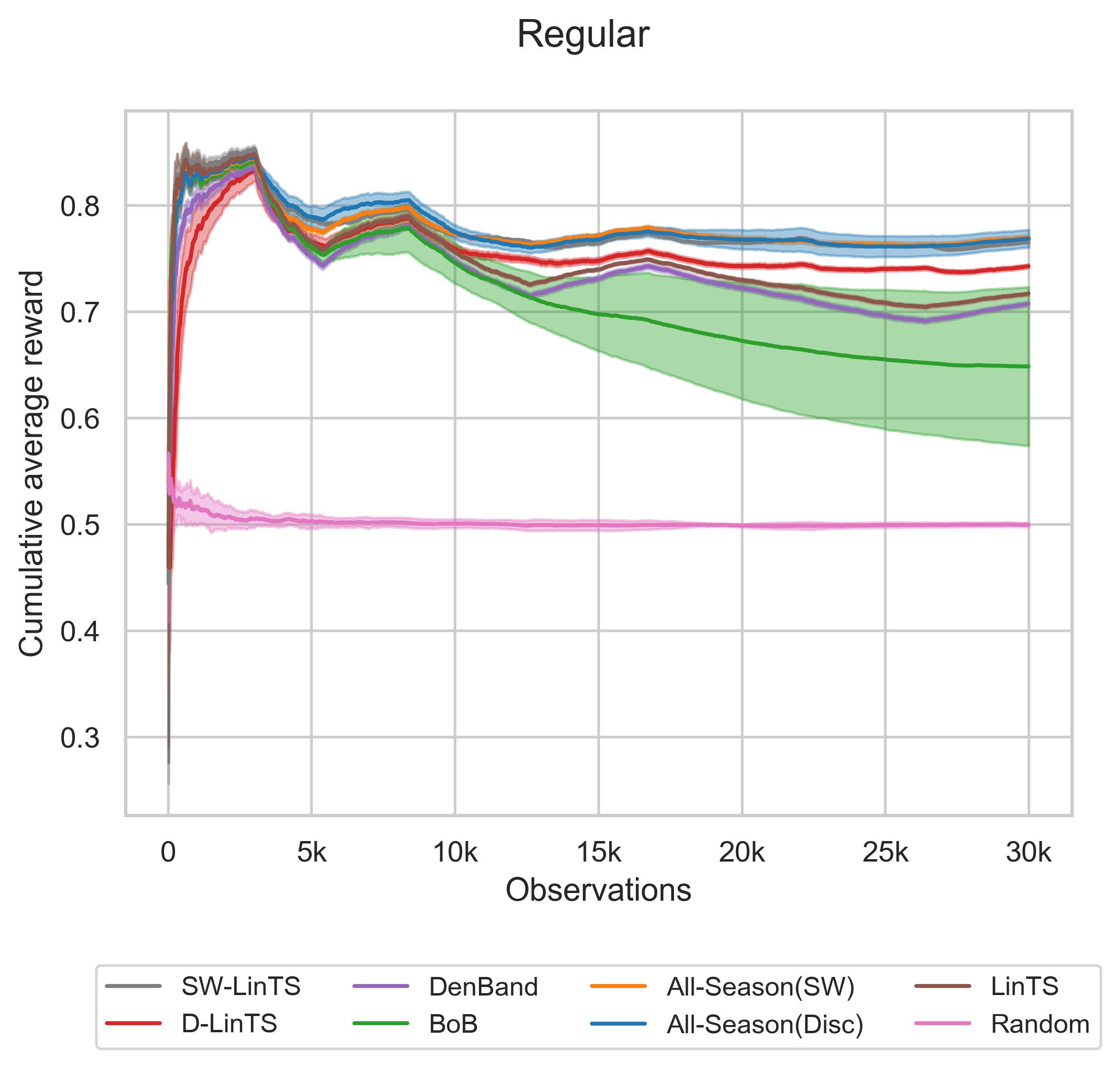} &
  \includegraphics[scale=.30]{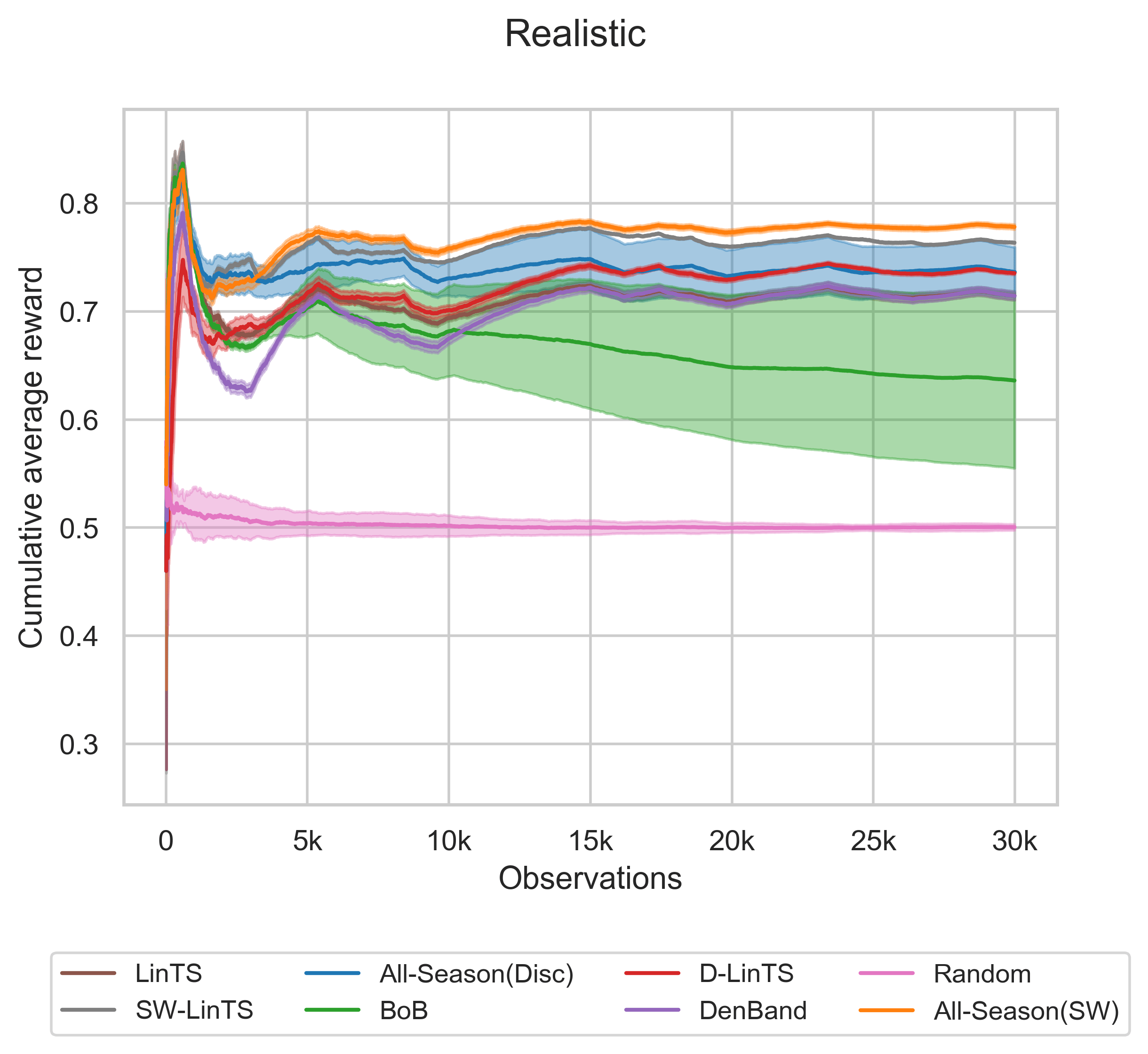} &
  \includegraphics[scale=.30]{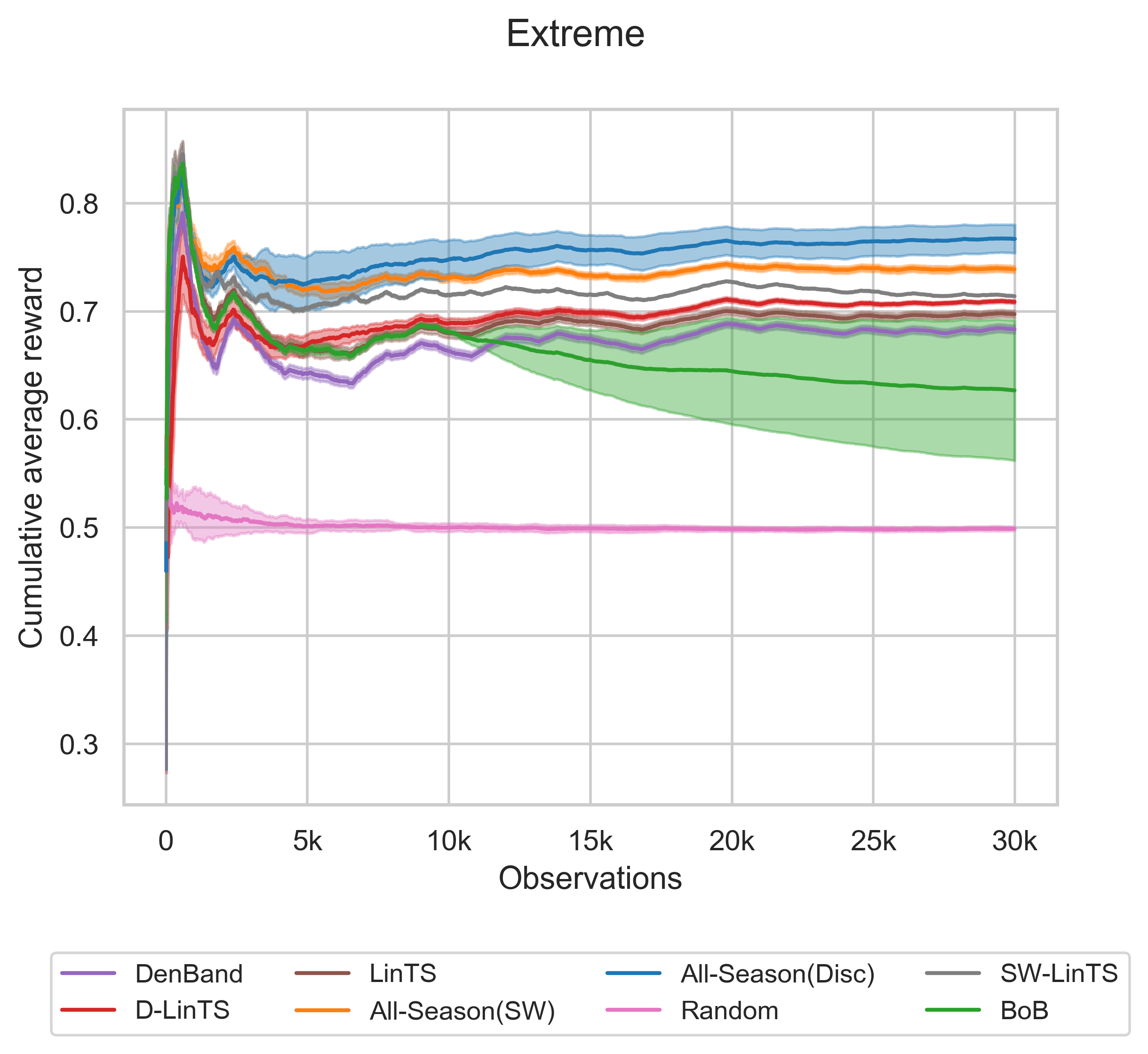}
\end{tabular}
\caption{Mnist two-arm: Regular, Realistic and Extreme settings} 
\label{fig:mnist2}

\begin{tabular}{ccc}
  \includegraphics[scale=.30]{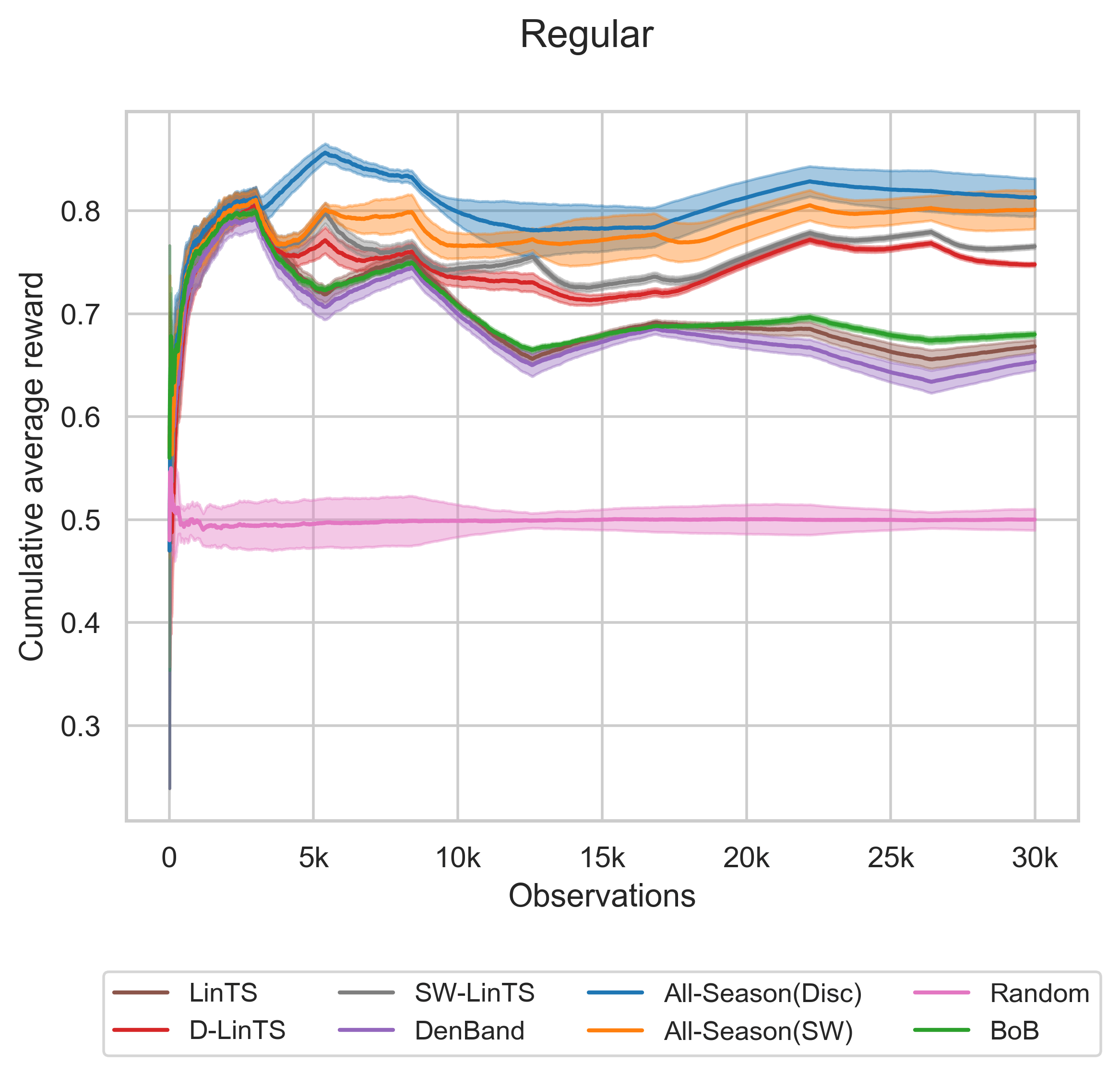} &
  \includegraphics[scale=.30]{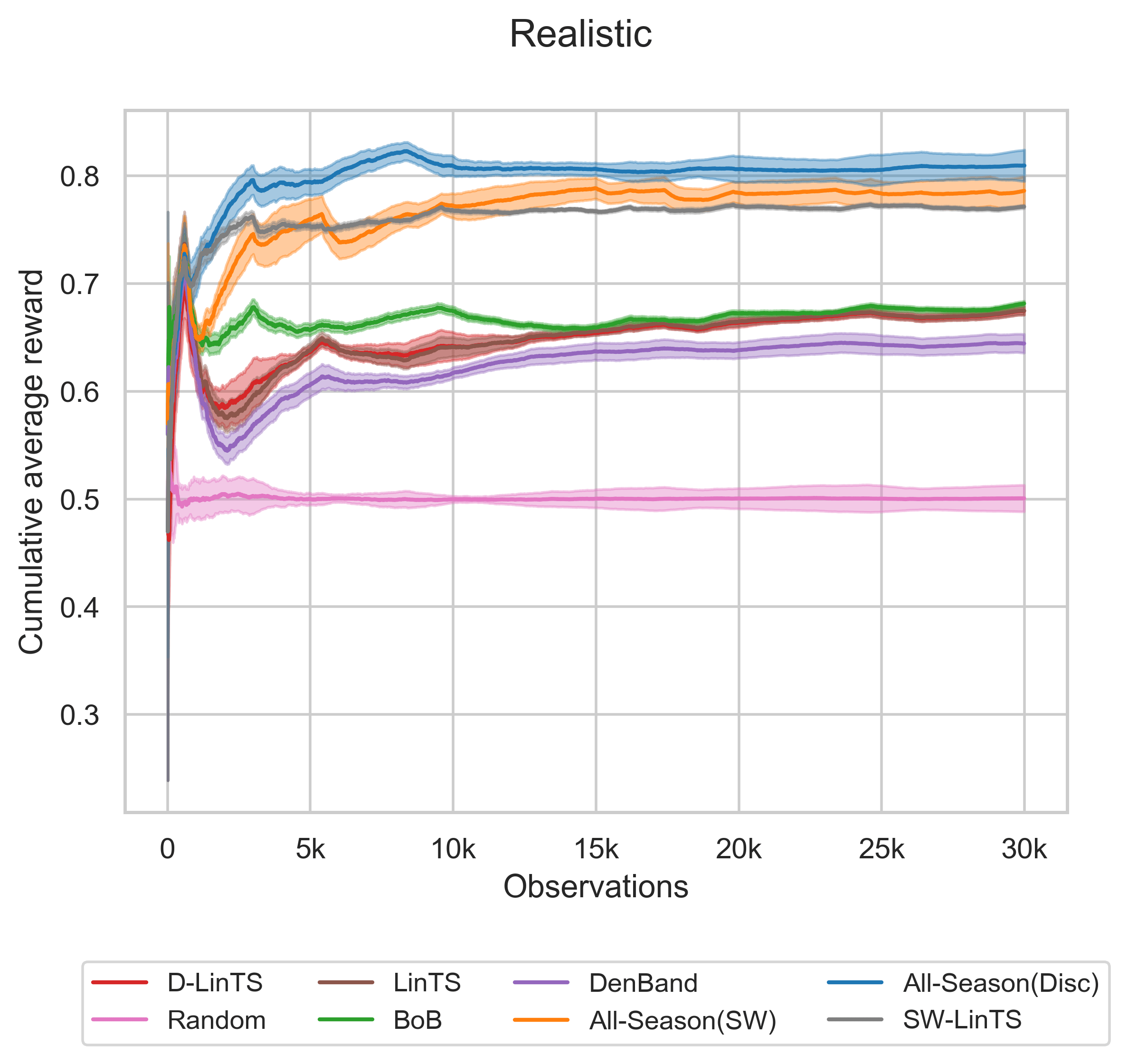} &
  \includegraphics[scale=.30]{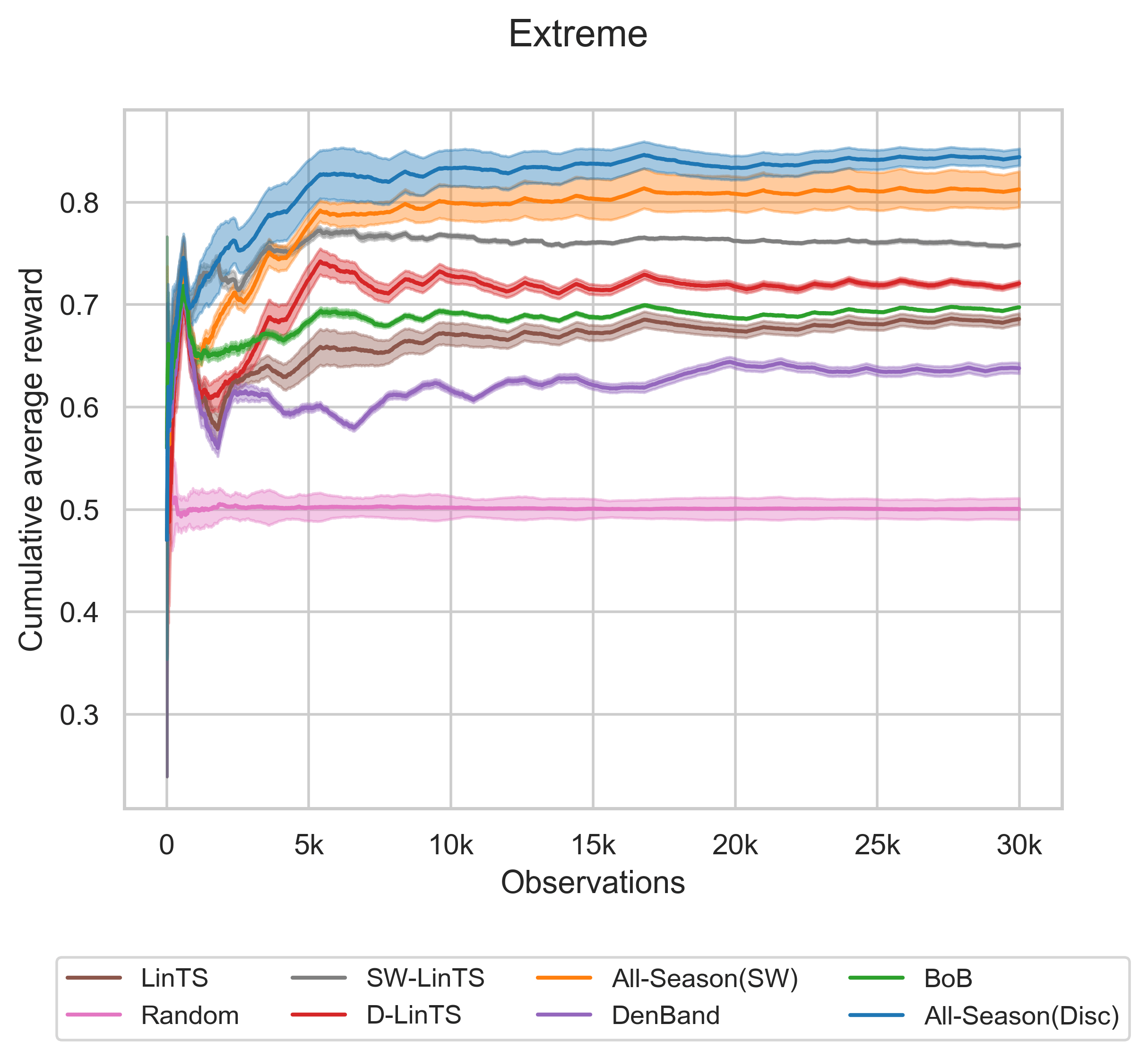}
  \end{tabular}
  \caption{Fashion-Mnist two-arm: Regular, Realistic and Extreme settings}
  \label{fig:fmnist2}

\begin{tabular}{ccc}
  \includegraphics[scale=.30]{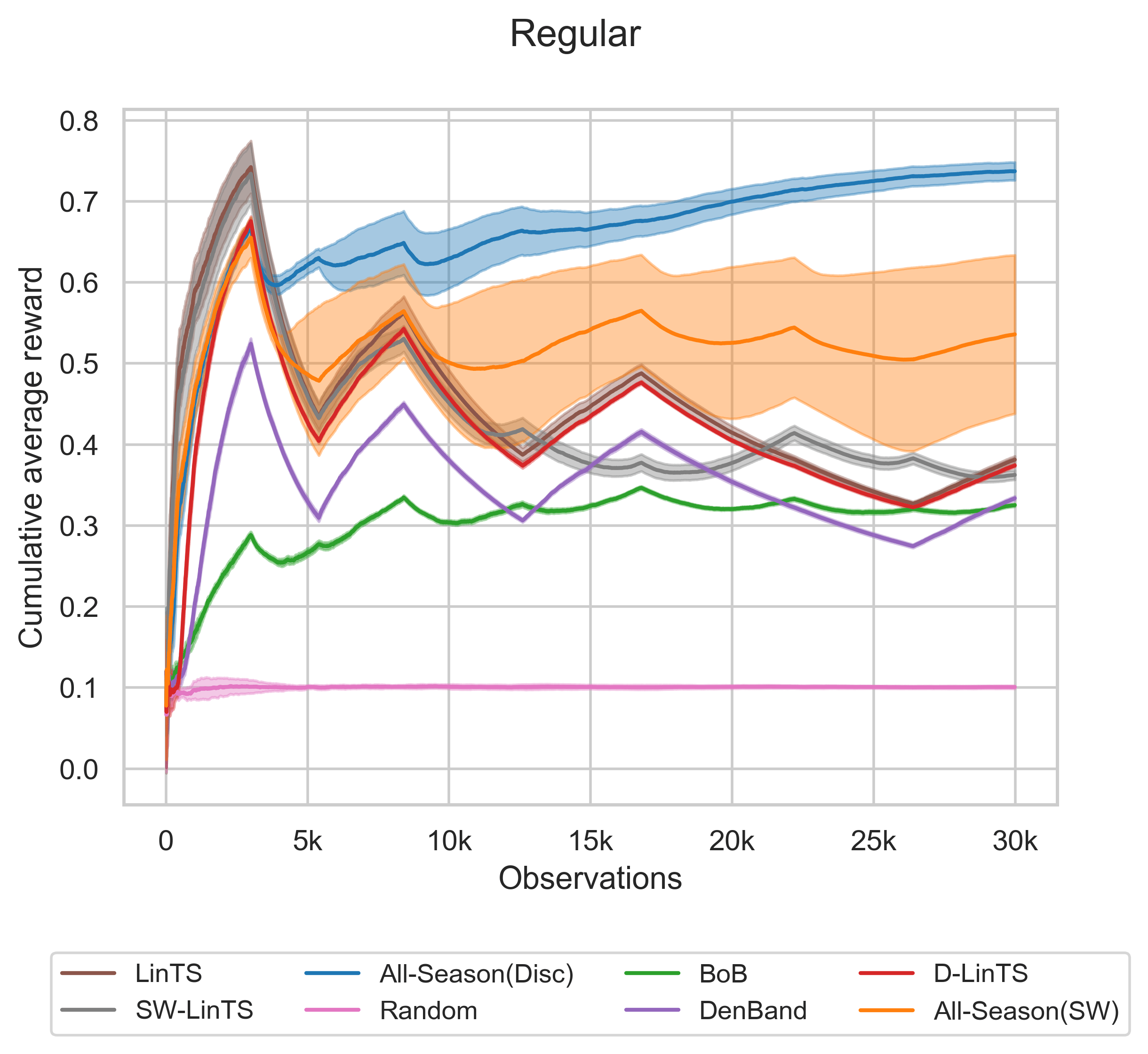} &
  \includegraphics[scale=.30]{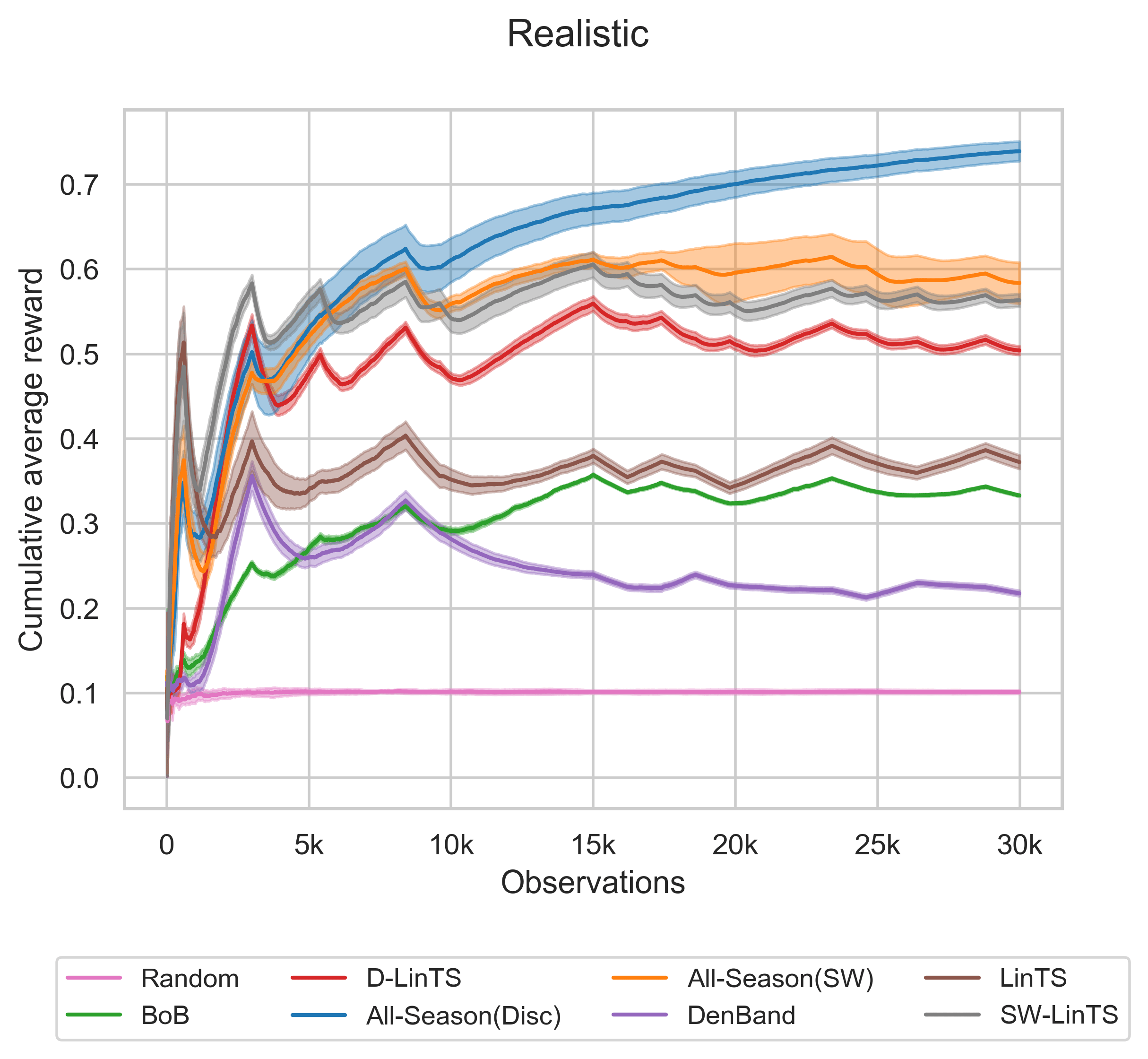} &
  \includegraphics[scale=.30]{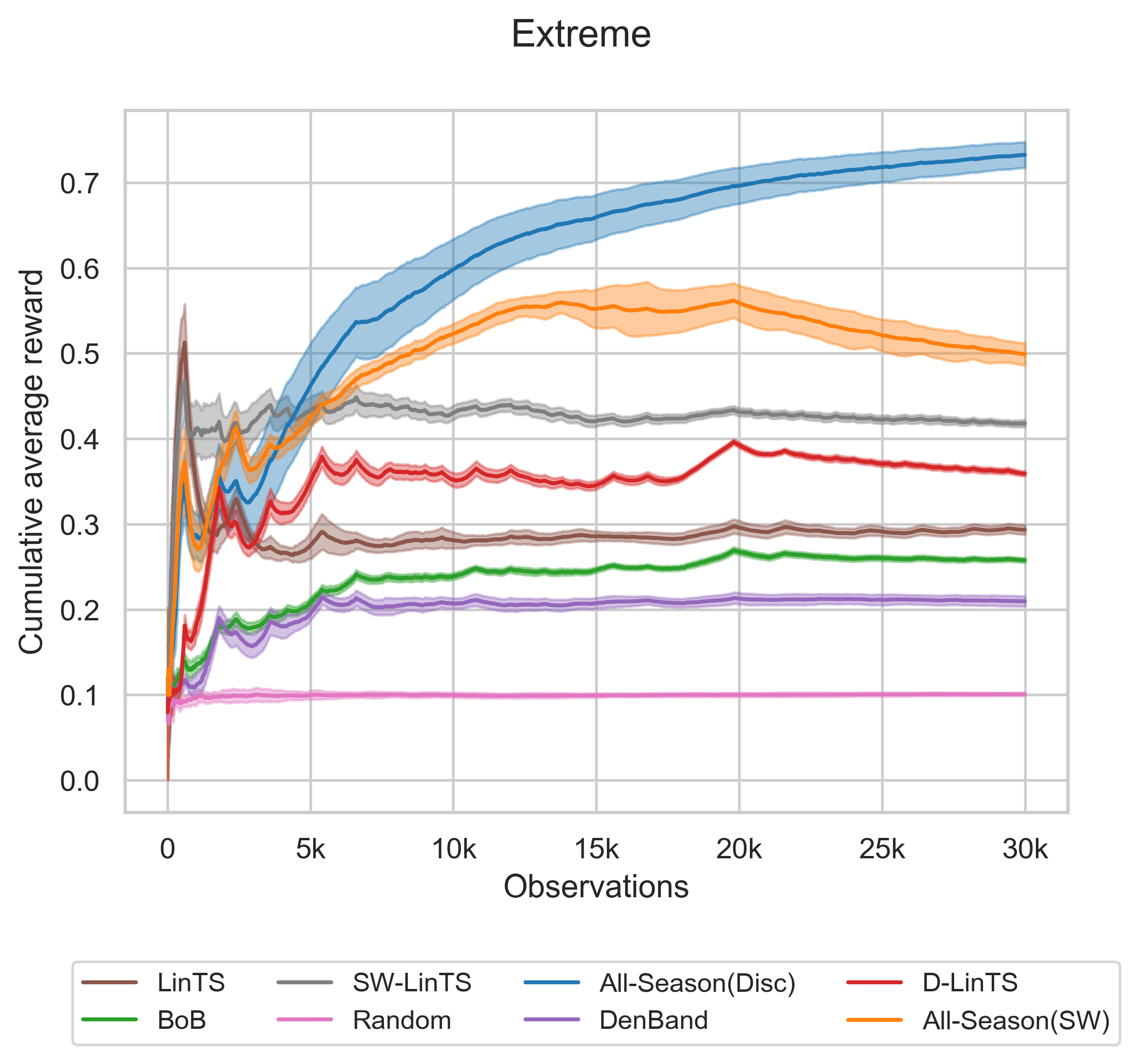}
  \end{tabular}
  \caption{Mnist digit-recognition: Regular, Realistic and Extreme settings}
  \label{fig:mnist10}

\begin{tabular}{ccc}
  \includegraphics[scale=.30]{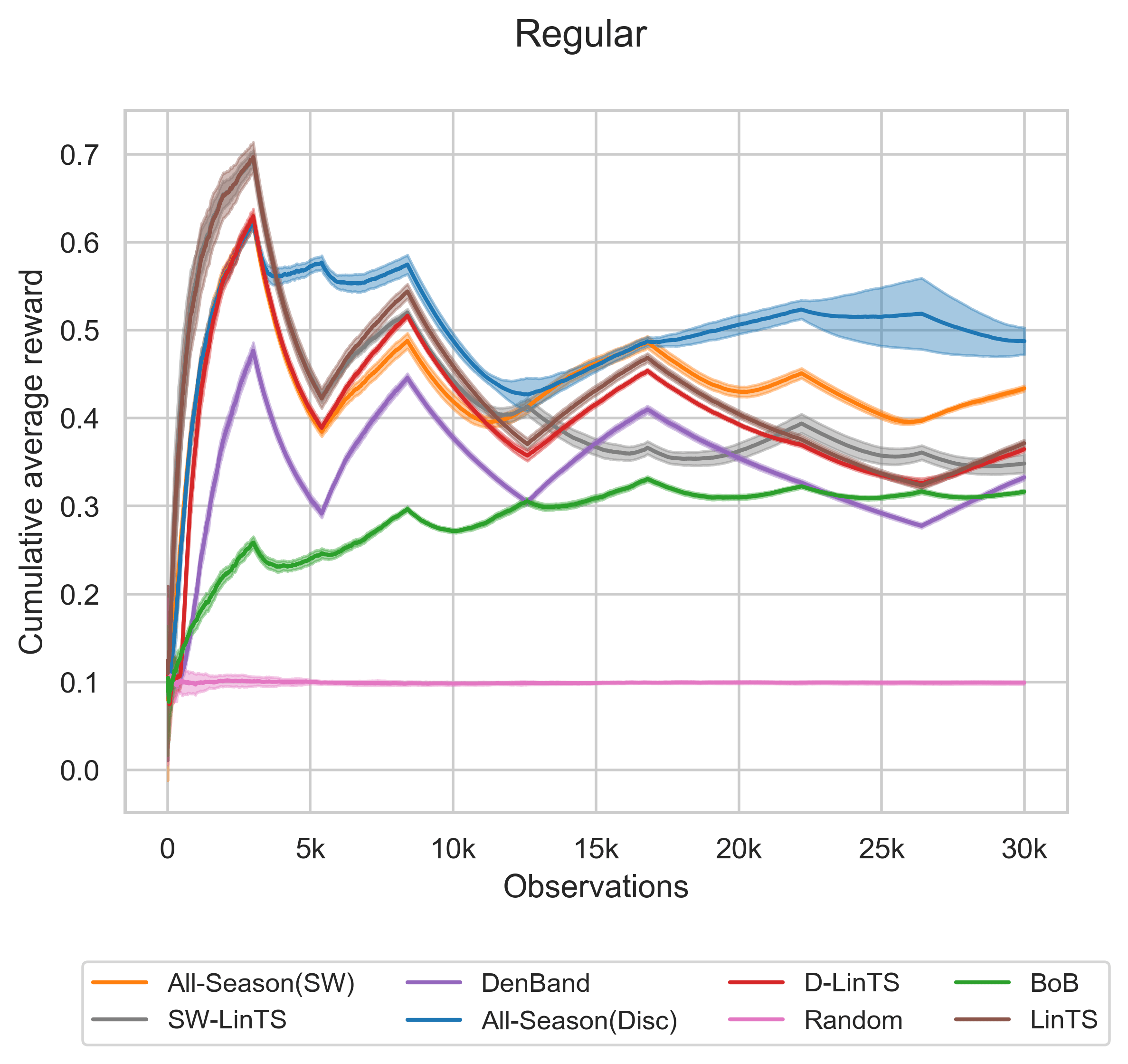} &
  \includegraphics[scale=.30]{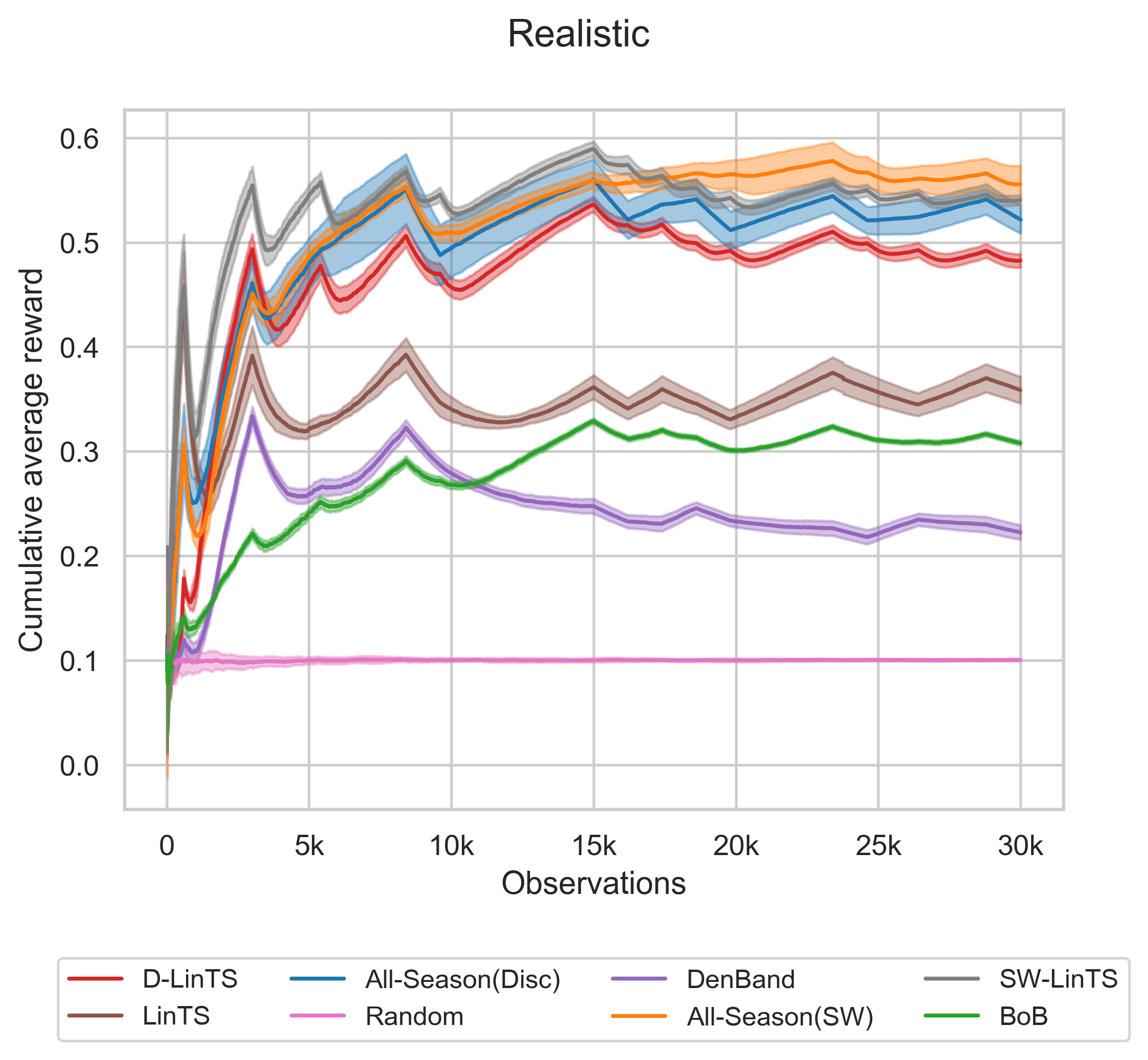} &
  \includegraphics[scale=.30]{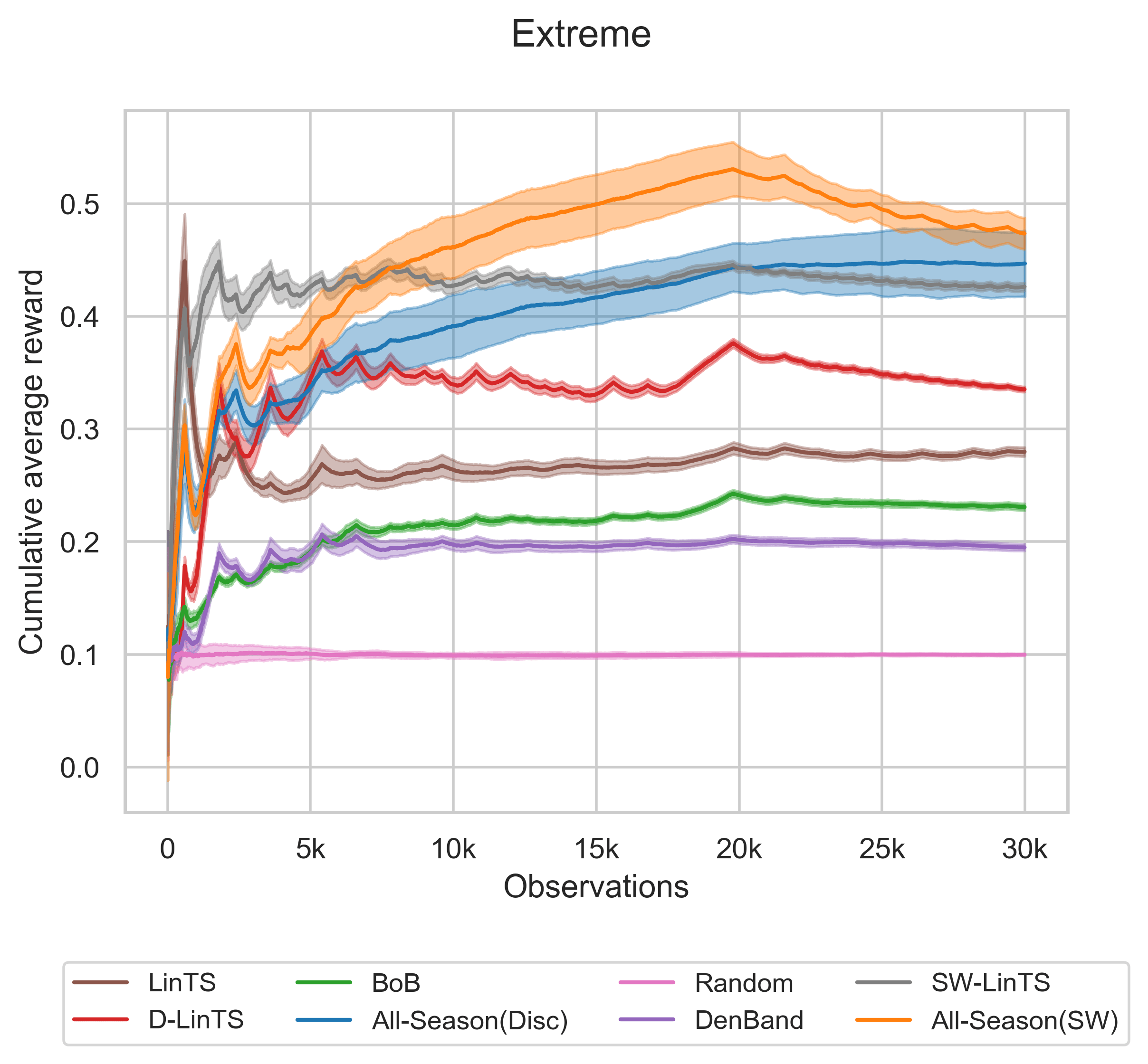}
  \end{tabular}
  \caption{Fashion-Mnist item-recognition: Regular, Realistic and Extreme settings}
  \label{fig:fmnist10}
\vspace{2cm}
\end{figure*}
\clearpage

\begin{table}[h!]
  \centering
  \begin{tabular}{rccc}
    \toprule
    & Regular & Realistic & Extreme \\
    \midrule
    \algo(Disc)    & 0.74 $\pm$ 0.01 & 0.74 $\pm$ 0.01 & 0.73 $\pm$ 0.02 \\
    \algo(SW)    & 0.54 $\pm$ 0.10 & 0.58 $\pm$ 0.02 & 0.50 $\pm$ 0.01  \\
    SW-LinTS   & 0.36 $\pm$ 0.00 & 0.56 $\pm$ 0.01 & 0.42 $\pm$ 0.00 \\
    D-LinTS    & 0.37 $\pm$ 0.00 & 0.50 $\pm$ 0.01 & 0.36  $\pm$ 0.00  \\
    DenBand    & 0.33 $\pm$ 0.00 & 0.22 $\pm$ 0.00 & 0.21 $\pm$ 0.01 \\
    BoB        & 0.32 $\pm$ 0.00 & 0.33 $\pm$ 0.00 & 0.26 $\pm$ 0.00  \\
    LinTS      & 0.38 $\pm$ 0.00 & 0.37 $\pm$ 0.01 & 0.29  $\pm$ 0.01 \\               
    random     & 0.10 $\pm$ 0.00 & 0.10 $\pm$ 0.00 & 0.10 $\pm$ 0.00 \\
    \bottomrule
  \end{tabular}
  \caption{\small Mnist digit-recognition: average reward $\pm$ std dev.}
  \label{table:mnist10}
\end{table}

\begin{table}[h!]
  \centering
  \begin{tabular}{rccc}
    \toprule
    & Regular & Realistic & Extreme \\
    \midrule
    \algo(Disc)    & 0.49 $\pm$ 0.02 & 0.52 $\pm$ 0.01 &  0.45 $\pm$ 0.03 \\
    \algo(SW)   & 0.43 $\pm$ 0.00 & 0.56 $\pm$ 0.00 & 0.47 $\pm$ 0.01 \\
    SW-LinTS   & 0.35 $\pm$ 0.01 & 0.54 $\pm$ 0.00 & 0.43 $\pm$ 0.00 \\
    D-LinTS    & 0.37 $\pm$ 0.00 & 0.48 $\pm$ 0.01 & 0.36 $\pm$ 0.00 \\
    DenBand    & 0.33 $\pm$ 0.00 & 0.22 $\pm$ 0.01 & 0.20 $\pm$ 0.00 \\
    BoB        & 0.32 $\pm$ 0.00 & 0.31 $\pm$ 0.00 & 0.23  $\pm$ 0.00 \\
    LinTS      & 0.37 $\pm$ 0.00 & 0.35 $\pm$ 0.01 & 0.28  $\pm$ 0.00 \\               
    random     & 0.10$\pm$ 0.00 & 0.10 $\pm$ 0.00 & 0.10 $\pm$ 0.00  \\
    \bottomrule
  \end{tabular}
  \caption{\small Fashion-Mnist item-recognition: average reward $\pm$ std dev.}
  \label{table:fmnist10}
\end{table}

\noindent simpler and more robust algorithm than its competitors, suitable to be maintained in industrial applications with little effort.

We suppose that there still room for improvement in dealing with the model misspecification problem and in 
selecting the models to be pruned. For the first problem, we believe that the General Bayes approach \cite{knoblauch2019generalized}, used in the definition of the posterior predictive weights, would be a more principled solution that can deliver a more robust algorithm, as already proven in online change-point detection for time series in \cite{knoblauch2018doubly}.
To improve the pruning strategies, we are considering techniques inspired by Bayesian Core Sets \cite{campbell2018bayesian}.

\nocite{*}
\bibliographystyle{ACM-Reference-Format}
\bibliography{biblio}

\end{document}